\documentclass[10pt,twocolumn,letterpaper]{article}

\usepackage{cvpr}
\usepackage{times}
\usepackage{epsfig}
\usepackage{graphicx}
\usepackage{amsmath}
\usepackage{amssymb}

\usepackage{algorithm,algpseudocode}
\usepackage{subfigure}
\usepackage{float}
\usepackage{color}
\usepackage{ulem}
\usepackage{ifthen}
\usepackage{array,multirow}
\usepackage{booktabs}
\usepackage{siunitx}
\usepackage{graphicx}
\usepackage{adjustbox}

\definecolor{green}{rgb}{0, 0.4, 0} 
\definecolor{orange}{rgb}{1, 0.5, 0} 	
\definecolor{mahogany}{rgb}{0.75, 0.25, 0.0}
\definecolor{purple}{rgb}{0.6, 0, 0.6}
\definecolor{purple}{rgb}{0.6, 0, 0.6}
\definecolor{darkgreen}{rgb}{0, 0.4, 0.4} 
\definecolor{frenchblue}{rgb}{0.0, 0.45, 0.73}
\definecolor{what_color}{rgb}{0.7, 0.4, 0.3}
\definecolor{spring_green}{rgb}{0, 1, 0.5}

\newboolean{revising}
\setboolean{revising}{false}
\ifthenelse{\boolean{revising}}
{
    \newcommand{\ignore}[1]{}

    \newcommand{\chengreplace}[2]{\textcolor{frenchblue}{#2}}

} {
    \newcommand{\ignore}[1]{}

    \newcommand{\chengreplace}[2]{#2}

}

\newcommand{\cutsectionup}{\vspace*{-0.1in}}

\usepackage[pagebackref=true,breaklinks=true,letterpaper=true,colorlinks,bookmarks=false]{hyperref}

\newcommand*{\affaddr}[1]{\normalsize#1}
\newcommand*{\affmark}[1][*]{\normalsize\textsuperscript{#1}}
\newcommand*{\email}[1]{\footnotesize\texttt{#1}}

\cvprfinalcopy 

\def\cvprPaperID{700} 

\ifcvprfinal\fi
\begin{document}

\title{Cube Padding for Weakly-Supervised Saliency Prediction in 360$^{\circ}$ Videos}


\author{
\normalsize
Hsien-Tzu Cheng\affmark[1],
Chun-Hung Chao\affmark[1],
Jin-Dong Dong\affmark[1],
Hao-Kai Wen\affmark[2],
Tyng-Luh Liu\affmark[3],
Min Sun\affmark[1] \\
\affaddr{\affmark[1]National Tsing Hua University} \hspace{1.5mm} 
\affaddr{\affmark[2]Taiwan AI Labs} \hspace{1.5mm} 
\affaddr{\affmark[3]Academia Sinica} \\
\email{hsientzucheng@gapp.nthu.edu.tw} \hspace{1.5mm} \email{\{raul.c.chao, mark840205\}@gmail.com}\\
\email{hao.kai@ailabs.tw} \hspace{1.5mm}
\email{liutyng@iis.sinica.edu.tw} \hspace{1.5mm}
\email{sunmin@ee.nthu.edu.tw}
}
\maketitle

\begin{abstract}

Automatic saliency prediction in 360$^{\circ}$ videos is critical for viewpoint guidance applications (e.g., Facebook 360 Guide).
We propose a spatial-temporal network which is (1) weakly-supervised trained and (2) tailor-made for 360$^{\circ}$ viewing sphere. Note that most existing methods are less scalable since they rely on annotated saliency map for training.
Most importantly, they convert 360$^{\circ}$ sphere to 2D images (e.g., a single equirectangular image or multiple separate Normal Field-of-View (NFoV) images) which introduces distortion and image boundaries.
In contrast, we propose a simple and effective Cube Padding (CP) technique as follows.
Firstly, we render the 360$^{\circ}$ view on six faces of a cube using perspective projection. Thus, it introduces very little distortion. Then, we concatenate all six faces while utilizing the connectivity between faces on the cube for image padding (i.e., Cube Padding) in convolution, pooling, convolutional LSTM layers.
In this way, CP introduces no image boundary while being applicable to almost all Convolutional Neural Network (CNN) structures. To evaluate our method, we propose Wild-360, a new 360$^{\circ}$ video saliency dataset, containing challenging videos with saliency heatmap annotations. In experiments, our method outperforms baseline methods in both speed and quality.



\end{abstract}

\vspace{-4mm}
\section{Introduction}\label{sec.Intro}
\vspace{-3mm}
\normalem

\begin{figure}[t!]
\begin{center}
\includegraphics[width=1\linewidth]{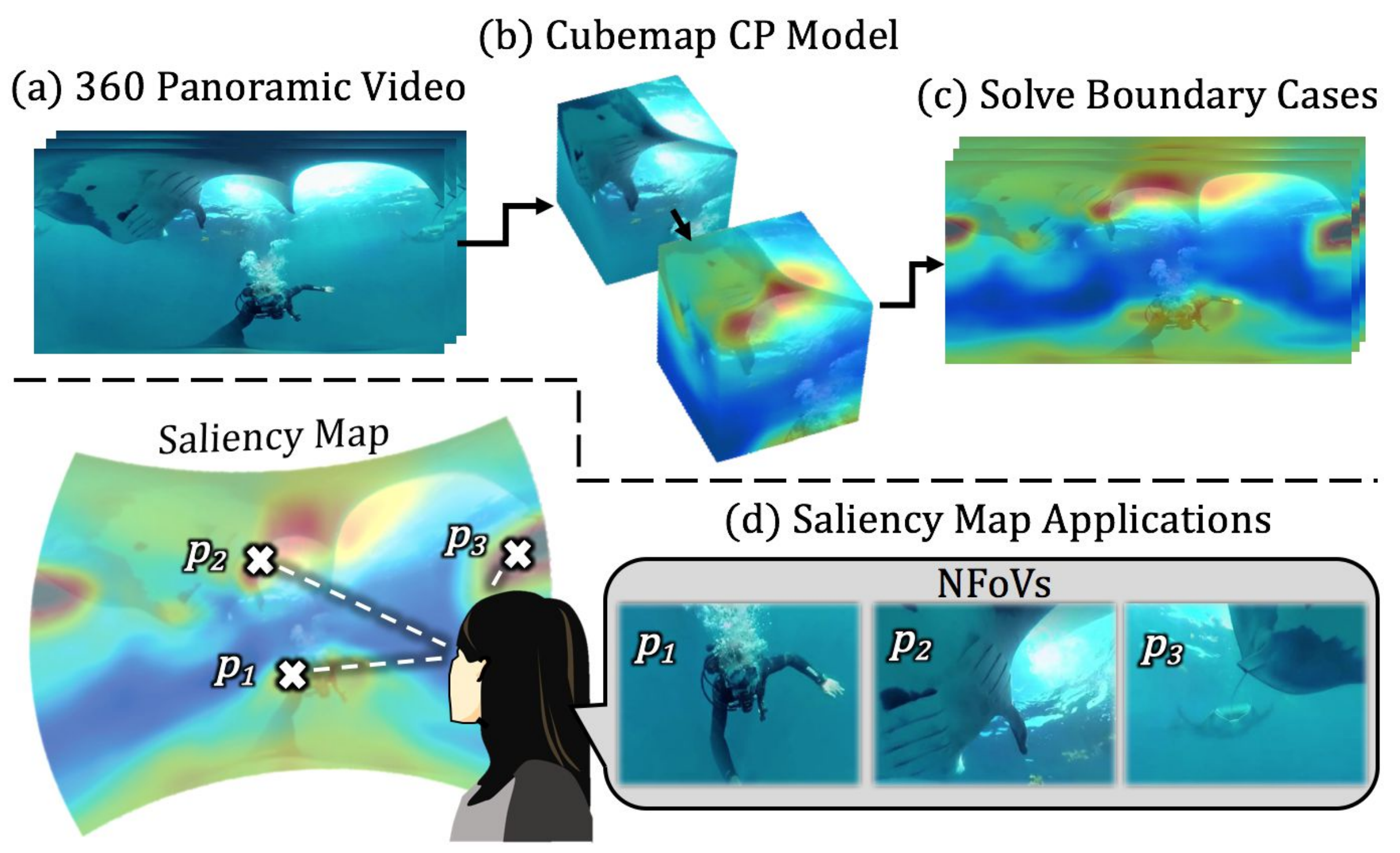}
\end{center}\vspace{-4mm}
\caption{
\small{Saliency prediction in a 360$^{\circ}$ video. Panel (a) shows a challenging frame in equirectangular projection with two marine creatures. One is near the north polar and the other is near the horizontal boundary. Panel (b) shows that Cubemap projection with Cube Padding (CP) mitigate distortion and cuts at image boundaries. As a result, we predict high-quality saliency map on the Cubemap. In panel (c), when visualizing our predicted saliency map on equirectangular, both marine creatures are recalled. In panel (d), desirable Normal Field of Views (NFoVs) are obtained from high-quality saliency map.}
} 
\label{fig.teaser}
\end{figure}

The power of 360$^{\circ}$ camera is to capture the entire viewing sphere (referred to as sphere for simplicity) surrounding its optical center, providing a complete picture of the visual world. This ability goes beyond the traditional perspective camera and the human visual system which both have a limited Field of View (FoV).
Videos captured using 360$^{\circ}$ camera (referred to as 360$^{\circ}$ videos) are expected to have a great impact in domains like virtual reality (VR), autonomous robots, surveillance systems in the near future. For now, 360$^{\circ}$ videos already gained its popularity thanks to low-cost hardware on the market, and supports of video streaming on YouTube and Facebook.

Despite the immersive experience and complete viewpoint selection freedom provided by 360$^{\circ}$ videos, many works recently show that it is important to guide viewers' attention. 
\cite{hu2017deep, lai2017semantic, supano2vid, su2017making} focus on selecting the optimal viewing trajectory in a 360$^\circ$ video so that viewers can watch the video in Normal FoV (NFoV). \cite{OutsideIn,TellME360} focus on providing various visual guidance in VR display so that the viewers are aware of all salient regions. Most recently, Chou \etal.~\cite{ChouViewGround360} propose to guide viewers' attention according to the scripts in a narrated video such as a tour guide video. Yu \etal.~\cite{YuHighlight360} propose to generate a highlight video according to spatial-temporal saliency in a 360$^\circ$ video.
All methods above involve predicting or require the existence of spatial-temporal saliency map in a 360$^\circ$ video.


Existing methods face two challenges in order to predict saliency on 360$^{\circ}$ videos. Firstly, 360$^{\circ}$ videos capture the world in a wider variety of viewing angles compared to videos with an NFoV. Hence, existing image \cite{mit-saliency-benchmark,jiang2015salicon} or video \cite{DIEM} saliency datasets are not ideal for training saliency prediction model. One way to overcome this challenge is to collect saliency dataset directly on 360$^{\circ}$ videos. In this direction, Facebook~\cite{facebook_heatmap} and Youtube~\cite{youtube_heatmap} are collecting users' viewing history on 360$^{\circ}$ videos. However, there is a chicken and egg problem. Without the ability to predict saliency and provide attention guidance, users will only be interested in viewing some popular 360$^{\circ}$ videos. Moreover, these data are proprietary which are not publicly available to the research community.

Secondly, most existing methods~\cite{hu2017deep, lai2017semantic,supano2vid,su2017making} apply their techniques to process images with the equirectangular projection (referred to as equirectangular images). However, equirectangular images introduce image boundaries, and create distortion significantly at the top and bottom regions. Both of them lead to prediction artifacts and make the learning task harder. An alternative is to divide the 360$^{\circ}$ sphere into multiple but ``separate" perspective images. Although it avoids distortion, it will introduce more image boundaries. We can also divide the 360$^{\circ}$ sphere into multiple ``overlapping" perspective images. Then, we only take the saliency prediction in a center sub-region in order to combine all predictions onto the whole sphere. However, this will require many more perspective images and significantly slow down the prediction process.
Recently, Su and Grauman~\cite{SuNIPS} propose a new training procedure and a new network structure (i.e., spherical convolution) to overcome this challenge. We argue that a simpler modification on existing CNN can overcome this challenge.

In this work, we propose a spatial-temporal network 
consisting of a static model and a ConvLSTM module to form a temporal model. The static model is inspired by \cite{zhou2016learning} which computes a class-wise activation \chengreplace{``vector"}{map} per image. We remove the global average pooling and convert the last fully connected layer into a 1-by-1 convolutional layer to obtain a static feature per image. Note that this static model is weakly-supervised trained by only monocular image level supervision, i.e. without 360$^\circ$ videos. After that, static features at several timesteps are fed into a ConvLSTM \cite{xingjian2015convolutional} module to aggregate temporal information. 
Our temporal model is also designed to be trained in an unsupervised manner.
During training, our loss function mainly enforces temporal consistency on two consecutive predicted saliency maps given precomputed optical flow. Note that the ConvLSTM module and temporal consistency loss encourage the predicted saliency map to be temporally smooth and motion-aware.

Most importantly, our model is tailor-made for 360$^{\circ}$ videos (see Fig.~\ref{fig.teaser}). Firstly, we project the 360$^{\circ}$ sphere on six faces of a cube, which introduces very little distortion. Then, we concatenate all six faces as an input while utilizing the connectivity between faces on the cube for image padding (referred to as Cube Padding (CP)) in convolution, pooling, convolutional LSTM layers.
In this way, CP introduces no image boundary while utilizing existing CNN layers. 
To evaluate our method, we propose Wild-360, a new 360$^{\circ}$ video saliency dataset, containing challenging videos with saliency heatmap annotations. According to experimental results, our method outperforms all baselines in both speed and quality.

We summarize our contributions as below:

\noindent\textbf{1.} We propose an \textbf{weakly-supervised} trained spatial-temporal saliency prediction model. This ensures that our approach is scalable in overcoming large viewpoint variation in 360$^\circ$ videos. To the best of our knowledge, it is the first method to tackle the 360$^\circ$ video saliency map prediction in an weakly-supervised manner.

\noindent\textbf{2.} We introduce Cube Padding tailor-made for 360$^\circ$ videos to mitigate \textbf{distortion and image boundaries}. This module is fast, effective, and generally applicable to almost all existing CNN architectures.

\noindent\textbf{3.} We collect a \textbf{new Wild-360 dataset} with challenging 360 videos. One-third of our dataset is annotated with per-frame saliency heatmap for evaluation. Similar to \cite{facebook_heatmap,youtube_heatmap}, we collect heatmap by aggregating viewers' trajectories, consisting of 80 viewpoints per-frame.

\noindent\textbf{4.} Experimental results show that our method outperforms baseline methods both in \textbf{speed and quality}.

\vspace{-2mm}
\section{Related work}\label{sec.RW}
\vspace{-3mm}
\normalem

To better comprehend the proposed method and the potential contributions, we discuss the recent developments of relevant techniques, including saliency map prediction, localization via weakly-supervised or unsupervised learning, and 360$^\circ$ vision.

\noindent\textbf{Saliency map prediction.}
Predicting where humans look in an image has been a popular task in computer vision. \cite{liu2011learning,HarelKP06,AchantaHES09,wang2016learning,zhang2016exploiting,perazzi2012saliency} focus on detecting salient regions in images. \cite{Liu_2016_CVPR,Jetley_2016_CVPR,mlnet2016,pan2016shallow,pan2017salgan,Bruce_2016_CVPR,Wang_2016_CVPR,WangWLZR16,TangW16} employ deep learning to achieve much better results. For videos, \cite{MSali09,STSali08,MahadevanTPAMI,SeoJOV,DIEM,lee2015low} rely on low-level appearance and motion cues as inputs. In addition, \cite{Judd09,GofermanTPAMI,rudoy2013learning,MatheSminchisescuPAMI2015,GazePlus} consider information such as face, people, objects, \etc. However, all these approaches demand heavy saliency supervision while our method requires no manual saliency annotations.

\noindent\textbf{Weakly-supervised localization.} Recent techniques typically leverage the power of CNNs to localize the targets in an image, where the CNNs are only trained with image-level labels. The approach in \cite{oquab2015object} designs a Global Max Pooling (GMP) layer to carry out object localization by activating discriminative parts of objects. Subsequently, Zhou \etal \cite{zhou2016learning} propose Global Average Pooling (GAP) to achieve a much better result on activating the object regions. \cite{wang2017learning,durand2017wildcat,alameda2017viraliency}
instead consider using other pooling layers. Our method treats the deep features from the last convolutional layer, encoded with objectness clues, as saliency features for further processing. Having obtained the spatial saliency maps by selecting maximum per-pixel responses, we can then use these spatial heatmaps to learn or predict temporal saliency. More recently, Hsu \etal \cite{hsu12weakly} develop two coupled ConvNets, one for image-level classifier and the other for pixel-level generator. By designing a well-formulated loss function and top-down guidance from class labels, the generator is demonstrated to output saliency estimation of good quality.

\begin{figure*}[t!]\vspace{-5mm}
\begin{center}
\includegraphics[width=1\linewidth]{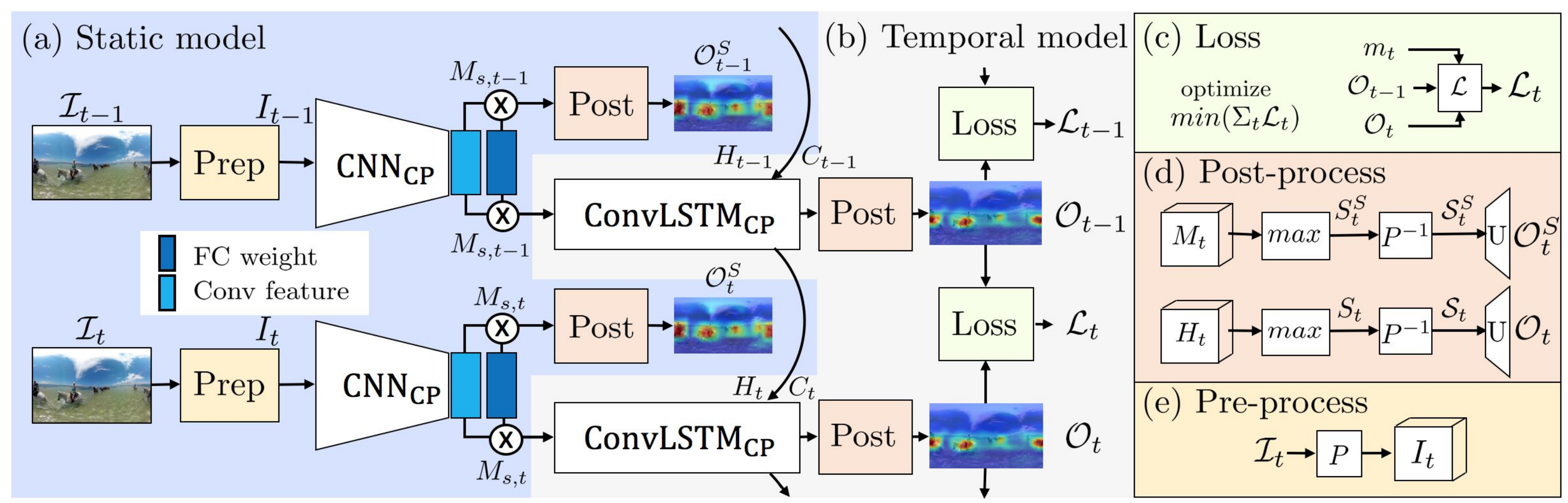}
\end{center}\vspace{-4mm}
\caption{\small Visualization of our system. Panel (a) shows our static model: (1) the pre-process to project an equirectangular image $\mathcal{I}$ to a cubemap image $I$, (2) the CNN with Cube Padding (CP) to extract a saliency feature $M_s$, (3) the post-process to convert $M_s$ into an equirectangular saliency map $\mathcal{O}^\mathcal{S}$.
Panel (b) shows our temporal model: (1) the convLSTM with CP to aggregate the saliency feature $M_s$ through time into $H$, (2) the post-process to convert $H$ into an equirectangular saliency map $\mathcal{O}$, (3) our self-supervised loss function to compute $\mathcal{L}_t$ given current $\mathcal{O}_t$ and previous $\mathcal{O}_{t-1}$. Panel (c) shows the total loss to be minimized. Panel (d) shows the post-process module including a max-pooling, inverse projection ($P^{-1}$), and upsampling (U). Panel (e) shows the pre-processing module with cubemap projection.
}
\label{fig.model}
\end{figure*}

%
\noindent\textbf{Unsupervised localization.}
One of the popular schemes for designing unsupervised deep-learning model is to train the underlying DNN with respect to the reconstruction loss. 
The reconstruction loss between an input and a warped image can be used for optical flow estimation \cite{zhu2017guided} and for single-view depth estimation \cite{garg2016unsupervised}. Turning now our attention to the unsupervised learning methods for video object segmentation, the two-stream neural network with visual memory by Tokmakov \etal \cite{tokmakov2017learning} is the current state-of-the-art for the benchmark, DAVIS \cite{pont20172017}. They generalize the popular two-stream architecture with ConvGRU \cite{ballas2015delving} to achieve the good performance. Although the network architecture of our method is not two-stream, it does explore the two-stream information sequentially, as shown in Figure~\ref{fig.model}. That is, the ConvLSTM \cite{xingjian2015convolutional} adopted in our approach is used to learn how to combine both spatial and temporal (including motion) information. While both \cite{tokmakov2017learning} and our work use self-supervision from video dynamics, we specifically focus on developing a general technique to solve the pole distortion and boundary discontinuity in processing 360$^\circ$ videos.

\noindent\textbf{360$^\circ$ Video.}
Different from the conventional, 360$^\circ$ videos bring in a whole distinct viewing experience with immersive content. The new way of recording yields, in essence, a spherical video that allows the users to choose the viewing directions for abundant scenarios as if they were in the center of filming environment.
In particular,  techniques related to virtual cinematography are introduced in  
\cite{supano2vid,su2017making,hu2017deep,lai2017semantic} to guide the user to make the FoV selection when viewing a 360$^\circ$ video. Nevertheless, such a strategy targets selecting a specific FoV and eliminates most of the rich content in a 360$^\circ$ video, while our proposed model generates a saliency map to activate multiple regions of interest. Indeed only a few attempts for estimating the saliency information in 360$^\circ$ videos have been made. The work by Monroy \etal~\cite{monroy2017salnet360} is the first to tackle this problem. To generate a saliency map for a 360$^\circ$ spherical patch, their method computes the corresponding 2D perspective image, and detect the saliency map using model pre-trained on SALICON dataset. Taking account of where the spherical patch is located at, the final result of saliency detection can be obtained by refining the 2D saliency map. However, defects due to the image boundaries are not explicitly handled.
In SALTINET \cite{Assens_2017_360Salient}, Assens \etal propose to predict scan-path of a 360$^\circ$ image with heavy manual annotations. Unlike our approach, these methods all require strong supervision. 

\noindent\textbf{Dataset.} 
One of the main contributions of our work is the effort to establish a new {\bf Wild-360} dataset. We thus briefly describe the current status of major collections relevant to (our) 360$^\circ$ video analysis. The MIT300 \cite{mit-saliency-benchmark} includes 300 benchmark images of indoor or outdoor scenes, collected from 39 observers using an eye tracker. It also comes with AUC-Judd and AUC-Borji evaluation metrics, which are adopted in our work. SALICON \cite{jiang2015salicon} has 10,000 annotations on MS COCO images, collected by a mouse-contingent multi-resolution paradigm based on neurophysiological and psychophysical studies of peripheral vision to simulate the natural viewing behavior of humans. The Densely-Annotated VIdeo Segmentation (DAVIS) \cite{pont20172017} is a new dataset with 50 high-resolution image sequences with all their frames annotated with pixel-level object masks. DIEM \cite{Mital2011} has, by  far, collected data from over 250 participants watching 85 different videos, and the fixations are reported with respect to the user's gaze. Finally, the Freiburg-Berkeley Motion Segmentation Dataset \cite{ochs2014segmentation} comprises a total of 720 frames, annotated with pixel-accurate segmentation annotation of moving objects. However, none of the datasets motioned above provides ground truth saliency map annotation on 360$^{\circ}$ videos to evaluate our proposed method.

\ULforem

\begin{figure}[t!]
\begin{center}
\includegraphics[width=0.8\linewidth]{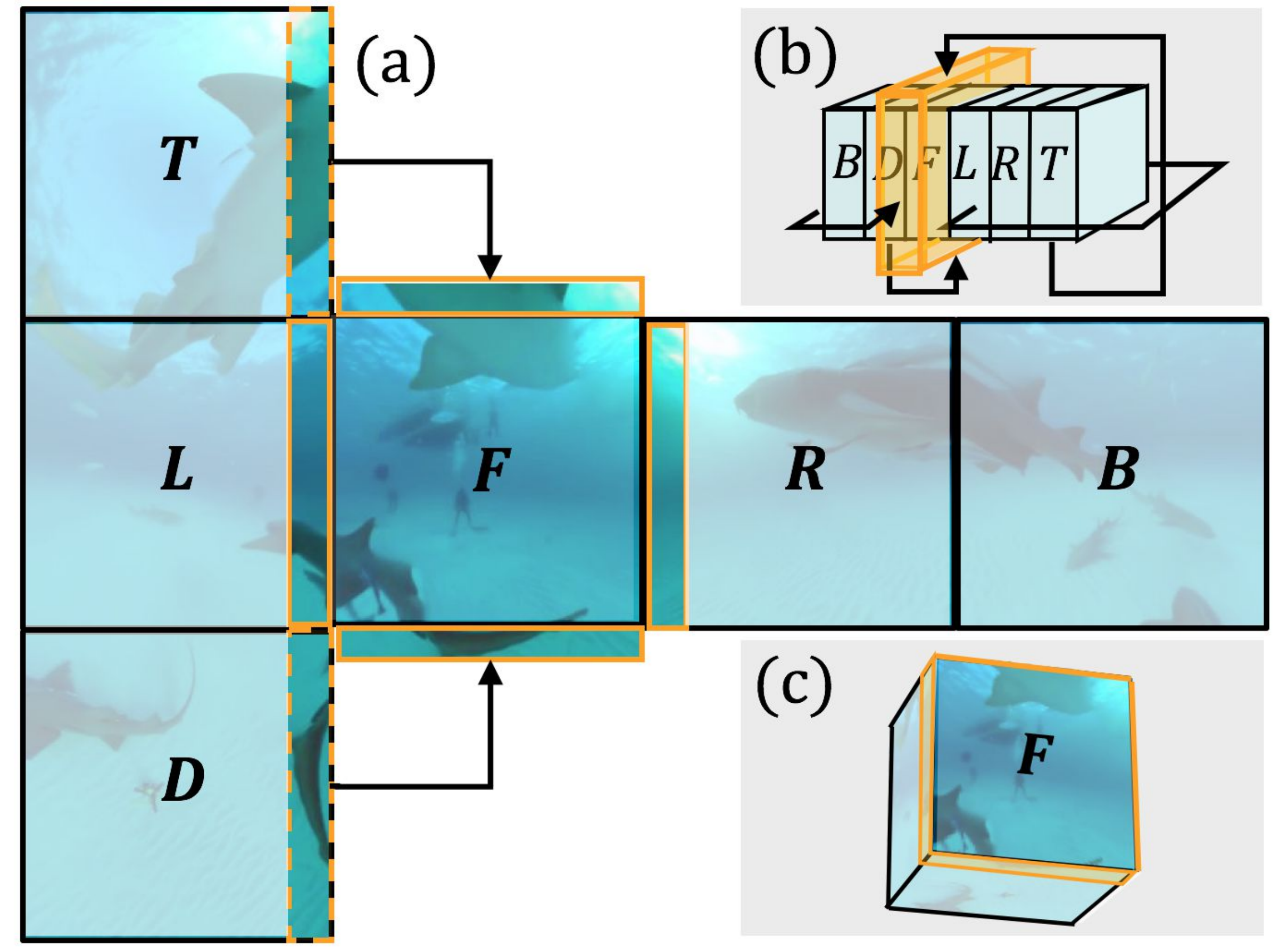}
\end{center}\vspace{-4mm}
\caption{\small{Illustration of Cube Padding (CP). In panel (a), we apply CP for the face $F$ which leverages information (in yellow rectangles) on face $T,L,R,D$ naturally rather than padding with zero values (i.e., zero padding). Panel (b) shows that this can be done in cubemap matric representation $M\in R^{6 \times c \times w \times w}$.
Panel (c) shows how to fold the faces back to a cube.}}
\label{fig.cp}
\end{figure}

\vspace{-2mm}
\section{Our method}\label{sec.Tech}
\vspace{-3mm}
In this section, we present our overall method as shown in Fig.~\ref{fig.model}, which consists of projection processes, static model, temporal model and loss functions. We describe Cube Padding and potential impacts in Sec.~\ref{sec.cp}, our static model in Sec.~\ref{sec.stat}, temporal model in Sec.~\ref{sec.temp}. Before that, we first introduce the various notations used in our formulation.

\vspace{-2mm}
\subsection{Notations}\label{sec.def}
\vspace{-3mm}
Given a 360$^\circ$ equirectangular 2D map $\mathcal{M} \in \mathbb{R}^{c \times q \times p}$ with the number of channels $c$, width $p$ and height $q$, we define a projection function $P$ to transform $\mathcal{M}$ to a cubemap representation $M \in \mathbb{R}^{6 \times c \times w \times w}$ with the edge length of the cube set to $w$. Specifically, $M$ is a stack of 6 faces $\{M^{B}, M^{D}, M^{F}, M^{L}, M^{R}, M^{T}\}$, where each face $M^j \in \mathbb{R}^{c \times w \times w}$, and $j\in \{B, D, F, L, R, T\}$ represents the $Back$, $Down$, $Front$, $Left$, $Right$, and $Top$ face, respectively. We can further inverse transform $M$ back to $\mathcal{M}$ by $\mathcal{M} = P^{-1}(M)$. Note that a RGB equirectangular image $\mathcal{I}$ is, in fact, a special 2D map where $c = 3$ and $I \in \mathbb{R}^{6 \times 3 \times w \times w}$ is a special cubemap with RGB value. For details of the projection function $P$ please refer to the supplementary material.

\vspace{-2mm}
\subsection{Cube padding}\label{sec.cp}
\vspace{-3mm}
Traditionally, Zero Padding (ZP) is applied at many layers in a Convolutional Neural Network (CNN) such as convolution and pooling. However, in our case, $M$ consists of 6 2D faces in a batch, observing the whole 360$^\circ$ viewing sphere. If we put $M$ to normal architecture with ZP in every single layer, the receptive field will be restricted inside each face, separating 360$^\circ$ contents into 6 non-connected fields. To solve this problem, we use Cube Padding (CP) to enable neurons to see across multiple faces by the interconnection between different faces in $M$. For an input $M$, CP takes the adjacent regions from the neighbor faces and concatenate them to the target face to produce a padded feature map. Fig.~\ref{fig.cp} illustrates a case of target face $M^{F}$ which is adjacent with $M^{R}, M^{T}, M^{L}$ and $M^{D}$. CP then simply considers the corresponding pads as shown in yellow patches in Fig.~\ref{fig.cp} outside $M^{F}$, where these pads are concatenated with $M^{F}$. Panel (a) in Fig.~\ref{fig.cp} illustrates that the yellow CP patch on the cubemap in 3D is visually similar to padding on sphere. Panel (b) shows the padding directions of $M^{F}$ in $M$ batch.

\begin{figure}[t!]\vspace{-5mm}
\begin{center}
\includegraphics[width=0.8\linewidth]{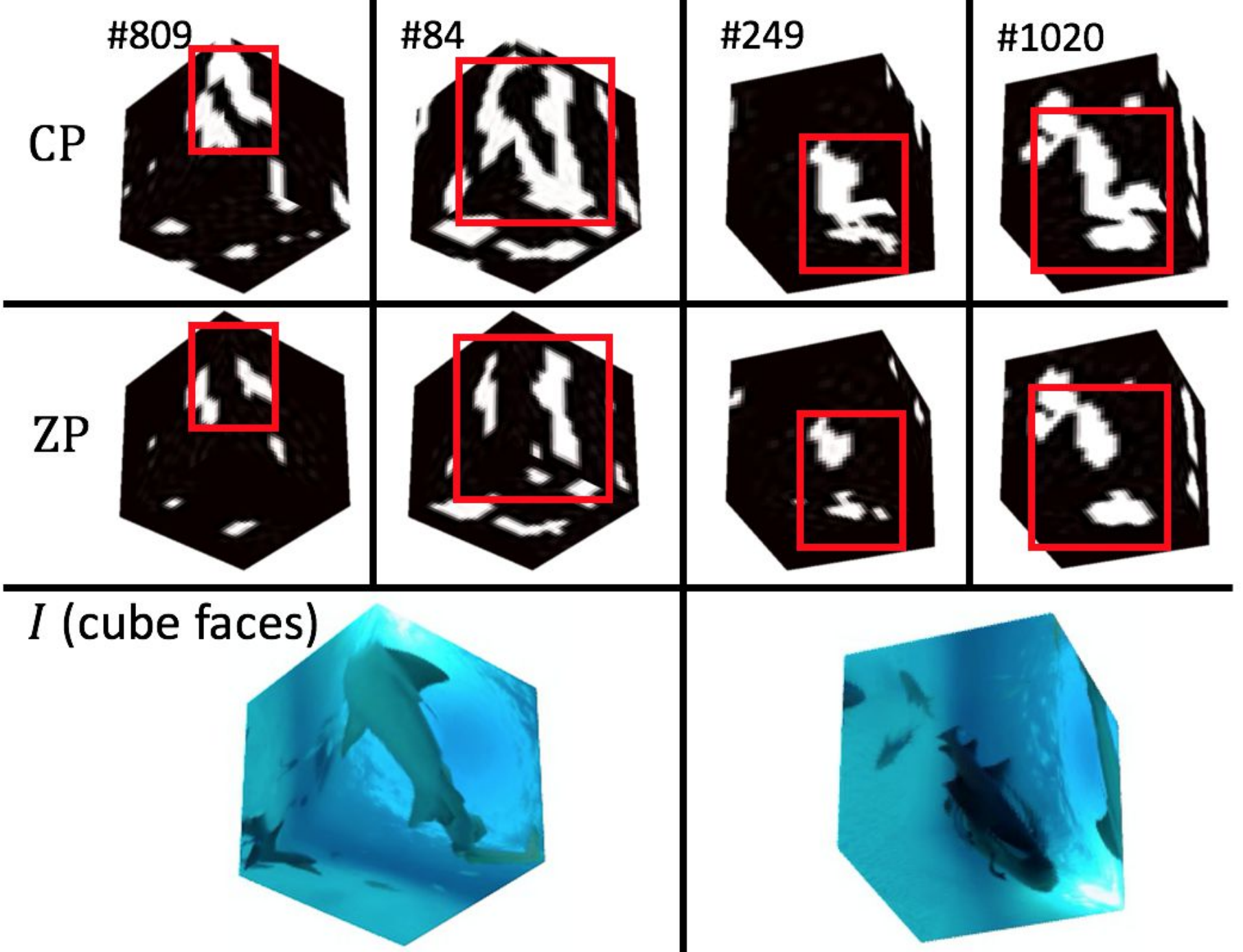}
\end{center}\vspace{-4mm}
\caption{\small{Feature map visualization from VGG Conv5\_3 layer. When Cube Padding (CP) is used (the first row), the response continuous through the face boundaries. However, when Zero Padding (ZP) is used (the second row), the responses near the boundaries vanished since each face is processed locally and separately. The last row shows the corresponding cubemap images containing several marine creatures across face boundaries.}}
\label{fig.deepvis}
\end{figure}

Although the padding size of CP is usually small, e.g. only 1 pixel for kernel size=3 and stride=1, by propagating $M$ through multiple layers incorporated with CP, the receptive field will gradually become large enough to cover contents across nearby faces. Fig.~\ref{fig.deepvis} illustrates some responses of deep features from CP and ZP. While ZP fails to have responses near the face boundaries, CP enables our model to recognize patterns of an object across faces.

To sum up, Cube Padding (CP) has following advantages: (1) applicable to most kinds of layers in CNN (2) the CP generated features are trainable to learn 360$^\circ$ spatial correlation across multiple cube faces, (3) CP preserves the receptive field of neurons across 360$^\circ$ content without the need for additional resolution.

\vspace{-2mm}
\subsection{Static model}\label{sec.stat}
\vspace{-3mm}
For each frame $\mathcal{I}$ of an input video sequence, our static model feeds preprocessed $I$ into the CNN. As shown in panel (a) of Fig.~\ref{fig.model}, CP module is incorporated in every convolutional and pooling layers in our CNN. The static model output $M_{S}$ is obtained by multiplying the feature map $M_{\ell}$ generated from the last convolutional layer with the weight of the fully connected layer $W_{fc}$.
\begin{align}
M_{S} = M_{\ell} \ast W_{fc} \label{eq.weightedf}
\end{align}
\noindent where $M_{S} \in \mathbb{R} ^ {6 \times K \times w \times w}$, $M_{\ell} \in \mathbb{R}^{6 \times c \times w \times w}$, $\mathcal{W}_{fc} \in \mathbb{R}^{c \times K \times 1 \times 1}$, $c$ is the number of channels, $w$ is corresponding feature width, ``$\ast$" means the convolution operation and $K$ is the number of classes for a model pre-trained on a specific classification dataset. To generate a static saliency map $S$, we simply pixel-wisely select the maximum value in $M_{S}$ along the class dimension.
\small
\begin{align}
S^{j}(x, y) = \max\limits_{k}\, \{M^{j}_{S}(k, x, y)\}~;
\forall j \in \{B, D, F, L, R, T\}~, \label{eq.cam}
\end{align}
\normalsize
where $S^{j}(x, y)$ is the saliency score at location $(x, y)$ of cube face $j$, and the saliency map in equirectangular projection $\mathcal{S}$ can be obtained with  $\mathcal{S} = P^{-1}(S)$. To get the final equirectangular output, we upsample $\mathcal{S}$ to $\mathcal{O}$ as shown in Fig.~\ref{fig.model} panel (d).

\vspace{-2.5mm}
\subsection{Temporal model}\label{sec.temp}
\vspace{-3mm}
\noindent\textbf{Convolutional LSTM.} 
Motivated by studies \cite{pratt2010s,meyerhoff2014interobject,meyerhoff2014perceptual}, human beings tend to put their attention on moving objects and changing scenes rather than static, we design our temporal model to capture dynamic saliency in a video sequence. As shown in the light gray block in Fig.~\ref{fig.model}, we use ConvLSTM as our temporal model, a recurrent model for spatio-temporal sequence modeling using 2D-grid convolution to leverage the spatial correlations in input data, which has been successfully applied to precipitation nowcasting \cite{xingjian2015convolutional} task. The ConvLSTM equations are given by
\small
\begin{align}
i_t ={} &\sigma(W_{xi} \ast M_{S,t} + W_{hi} \ast H_{t-1} + W_{ci} \circ C_{t-1} + b_i) \nonumber \\
f_t ={} &\sigma(W_{xf} \ast M_{S,t} + W_{hf} \ast H_{t-1} + W_{cf} \circ C_{t-1} + b_f)\nonumber \\
g_t ={} & \tanh(W_{xc} \ast X_t + W_{hc} \ast H_{t-1} + b_c) \nonumber \\
C_t ={} &i_t \circ g_t + f_t \circ C_{t-1} \nonumber \\
o_t ={} &\sigma(W_{xo} \ast M_t + W_{ho} \ast H_{t-1} + W_{co} \circ C_{t} + b_o)\nonumber \\
H_t ={} &o_t \circ tanh(C_t)~, \label{eq.ConvLSTM}
\end{align}
\normalsize
where $\circ$ denotes the element-wise multiplication, $\sigma(\cdot)$ is the sigmoid function, all $W_{*}$ and $b_{*}$ are model parameters to be learned, $i,f,o$ are the input, forget, and output control signals with value $[0,1]$, $g$ is the transformed input signal with value $[-1,-1]$, $C$ is the memory cell value, $H\in R^{6\times K \times w \times w}$ is the hidden representation as both the output and the recurrent input, $M_{S}$ is the output of the static model (see Eq.~(\ref{eq.weightedf})), $t$ is the time index which can be used in subscript to indicate timesteps.


We generate saliency map from $H_t$ equivalent to Eq.~(\ref{eq.cam}).
\small
\begin{align}
S^{j}_{t}(x, y) = \max\limits_{k}\, \{H^{j}_{t}(k, x, y)\}~;
\forall j \in \{B, D, F, L, R, T\}~, \label{eq.cam2}
\end{align}
\normalsize
where $S^{j}_t(x, y)$ is the generated saliency score at location $(x, y)$ of cube face $j$ at time step $t$. Similar to our static model, we upsample $\mathcal{S}$ to $\mathcal{O}$ to get the final equirectangular output.

\noindent\textbf{Temporal consistent loss.} 
Inspired by \cite{hsu12weakly,zhu2017guided,garg2016unsupervised} that model correlation between discrete images in an self-supervised manner by per-pixel displacement warping, smoothness regularization, etc., we design 3 loss functions to train our model and refine $O_t$ by temporal constraints: temporal reconstruction loss $\mathcal{L}^{recons}$, smoothness loss $\mathcal{L}^{smooth}$, and motion masking loss $\mathcal{L}^{motion}$. The total loss function of each time step $t$ can be formulated as:
\begin{align}
\mathcal{L}_t^{total} &= \lambda_r \mathcal{L}_t^{recons} + \lambda_s \mathcal{L}_t^{smooth} + \lambda_m \mathcal{L}_t^{motion}
\end{align}

In the following equations, i.e. Eqs.~(\ref{eq.temp})--(\ref{eq.motionmask}), $N$ stands for the number of pixels along spatial dimensions of one feature map, $O_t(p)$ is the output at pixel position $p$ at time step $t$, and $m$ is optical flow by \cite{weinzaepfel2013deepflow}.
$\mathcal{L}_t^{recons}$ is computed as the photometric error between the true current frame $O_{t}$ and the warped last frame $O_{t-1}(p+m)$:
\small
\begin{align}\label{eq.temp}
\mathcal{L}_t^{recons} &= \frac{1}{N} \sum^N|| O_{t}(p)-O_{t-1}(p+m)||^2
\end{align}
\normalsize

The reconstruction loss is formed by an assumption: the same pixel across different short-term time step should have a similar saliency score. This term helps to refine the saliency map to be more consistent in patches i.e. objects with similar motion patterns.
$\mathcal{L}_t^{smooth}$ is computed by the current frame and the last frame as:
\small
\begin{align}\label{eq.smooth}
\mathcal{L}_t^{smooth} &= \frac{1}{N} \sum^N || O_t(p)-O_{t-1}(p)||^2
\end{align}
\normalsize

The smoothness term is used to constrain the nearby frames to have a similar response without large changes. It also restrains the other 2 terms with motion included, since the flow could be noisy or drifting.
$\mathcal{L}_t^{motion}$ is used for motion masking:
\small
\begin{align}
\mathcal{L}_t^{motion} &= \frac{1}{N} \sum^N ||O_t(p)-O_t^m(p)||^2
\end{align}
\begin{equation}\label{eq.motionmask}
\begin{split}
O_t^m = \begin{cases}
0, & \mbox{if\;\;$|m(p)| \leq \epsilon$};\\
O_t(p), & \mbox{elsewhere}.
\end{cases}
\end{split}
\end{equation}
\normalsize
We set $\epsilon$ in Eq.~(\ref{eq.motionmask}) as a small margin to eliminate the pixel response where motion magnitude lowers than $\epsilon$. If a pattern in a video remains steady for several time steps, it is intuitively that the video saliency score of these non-moving pixels should be lower than changing patches.

For sequence length of ConvLSTM set to $Z$, the aggregated loss will be $\mathcal{L}^{total} = \sum^Z \mathcal{L}_t^{total}$. By optimizing our model with these loss functions jointly to $\mathcal{L}^{total}$ throughout the sequence, we can get the final saliency result by considering temporal patterns though $Z$ frames.



\vspace{-2mm}
\section{Dataset}\label{sec.Dataset}
\vspace{-3mm}


For the purpose of testing and benchmarking saliency prediction on 360$^{\circ}$ videos, a first and freshly collected dataset named Wild-360 is presented in our work. Wild-360 contains 85 360$^{\circ}$ video clips, totally about 55k frames. 60 clips within our dataset are for training and the rest 25 clips are for testing. All the clips are cleaned and trimmed from 45 raw videos obtained from YouTube. We manually select raw videos from keywords ``Nature'', ``Wildlife'', and ``Animals''; these keywords were selected in order to get videos with the following aspects: (i) sufficiently large number of search results of 360$^{\circ}$ video on YouTube, (ii) multiple salient objects in a single frame with diverse categories, (iii) dynamic contents inside the videos to appear in regions of any viewing angles including polar and borders.
The Wild-360 dataset is also designed to be diverse in object presence and free from the systematic bias. We rotate each testing video in both longitude and latitude angle to prevent the center-bias in ground truth saliency.


Recently, \cite{facebook_heatmap,youtube_heatmap} both announced to collect saliency heatmap of 360$^{\circ}$ videos by aggregating the viewers' trajectories during manipulation with view ports. To adopt the similar approach, but also giving the global perspective to viewers to easily capture multiple salient regions without missing hot spots, we adopt HumanEdit interface from \cite{supano2vid}. HumanEdit, as the Wild-360 labeling platform, encourages labelers to directly record attention trajectories based on their intuition. 30 labelers were recruited to label the videos in testing set, and they were asked to annotate from several viewing angles $\psi \in \lbrace 0^{\circ}, 90^{\circ}, 180^{\circ} \rbrace$. Therefore, there are about totally 80 viewpoints in a single frame. During annotation, videos and 3 rotation angles are shuffled to avoid order effect. In this setting, various positions could be marked as salient regions. Similar to \cite{tatler07}, we further apply Gaussian mask to every viewpoint to get aggregated saliency heatmap. Typical frames with ground truth heatmap (GT) are shown in the supplementary material. In order to foster future research related to saliency prediction in 360 videos, we plan to release the dataset, once the paper is published.

\vspace{-2mm}
\section{Experiments}\label{sec.Exp}
\vspace{-3mm}
We compare our saliency prediction accuracy and speed performance with many baseline methods.
In the following, we first give the implementation details. Then, we describe the baseline methods and evaluation metric. Finally, we report the performance comparison.

\vspace{-2mm}
\subsection{Implementation details}\label{sec.imp}
\vspace{-3mm}

We use ResNet-50~\cite{residualnet} and VGG-16~\cite{vgg} pretrained on ImageNet~\cite{deng2009imagenet} to construct our static model. For temporal model, we set $Z$ of ConvLSTM to 5 and train it for 1 epoch with ADAM optimizer and learning rate $10^{-6}$. We set the hyperparameters of temporal loss function to balance each term for steady loss decay. We set $\lambda_r=0.1$, $\lambda_s=0.7$, $\lambda_t=0.001$.
To measure the computational cost and quality performance of different settings, we set $w = 0.25p$, where $w$ and $p$ is the width of the cubemap and equirectangular image respectively. Moreover, the width of the equirectangular is 2 times the height of the equirectangular image, $q = 0.5p$. This setting is equivalent to \cite{facebook_cubemap} and fixes the total area ratio between cubemap and equirectangular image to 0.75. We implement all the padding mechanism rather than using built-in backend padding for fair comparison.

To generate ground truth saliency map of Wild-360, referring to \cite{tatler07} and heatmap providers \cite{facebook_heatmap}, the saliency distribution was modeled by aggregating viewpoint-centered Gaussian kernels. We set $\sigma=5$ to lay Gaussian inside the NFoV for invisible boundaries. To avoid the criterion being too loose, only locations on heatmap with value larger than $\mu+3\sigma$ were considered ``salient" when creating the binary mask for the saliency evaluation metrics, e.g. AUC.

\vspace{-2mm}
\subsection{Baseline methods}\label{sec.base}
\vspace{-3mm}

\noindent\textbf{Our variants.}

\noindent\textsl{\underline{Equirectangular (EQUI)}} --- We directly feed each equirectangular image in a $360^{\circ}$ video to our static model.

\noindent\textsl{\underline{Cubemap+ZP (Cubemap)}} --- As Sec.~\ref{sec.cp} mentioned, our static model takes the six faces of the cube as an input to generate the saliency map. However, unlike CP, Zero Padding (ZP) is used by the network operations, i.e. convolution, pooling, which causes the loss of the continuity of the cube faces.

\noindent\textsl{\underline{Overlap Cubemap+ZP (Overlap)}} --- We set FoV $= 120^\circ$ so that each face overlaps with each other by $15^\circ$. This variant can be seen as a simplified version of CP that process with larger resolution to cover the content near the border of each cube face. Note that this variant has no interconnection between faces, which means only ZP is used.

\noindent\textsl{\underline{EQUI + ConvLSTM}} --- We feed each equirectangular image to our temporal model to measure how much better the temporal model improves over static model.

\noindent\textbf{Existing methods.}

\noindent\textsl{\underline{Motion Magnitude}} --- As Sec.~\ref{sec.temp} mentioned, most salient regions in our videos are non-stationary. Hence we directly use the normalized magnitude of \cite{weinzaepfel2013deepflow} as saliency map to see how much motion clue contributes to video saliency.

\noindent\textsl{\underline{Consistent Video Saliency}} --- \cite{wang2015consistent} detects salient regions in spatio-temporal structure based on the gradient flow and energy optimization. It was the state-of-the-art video saliency detection methods on SegTrack \cite{TsaiBMVC10} and FBMS \cite{ochs2014segmentation}.

\noindent\textsl{\underline{SalGAN}} --- \cite{pan2017salgan} proposed a Generative Adversarial Network (GAN) to generate saliency map prediction. SalGAN is the current state-of-the-art model on well-known traditional 2D saliency dataset SALICON \cite{jiang2015salicon} and MIT300 \cite{mit-saliency-benchmark}. Note that this work focuses on saliency prediction on single image and needs ground truth annotations to do supervised learning. Hence, it cannot be trained on our dataset.

\vspace{-2mm}
\subsection{Computational efficiency}\label{sec.speed}
\vspace{-3mm}

\begin{figure}[t!]\vspace{-4mm}
\begin{center}
\includegraphics[width=1\linewidth]{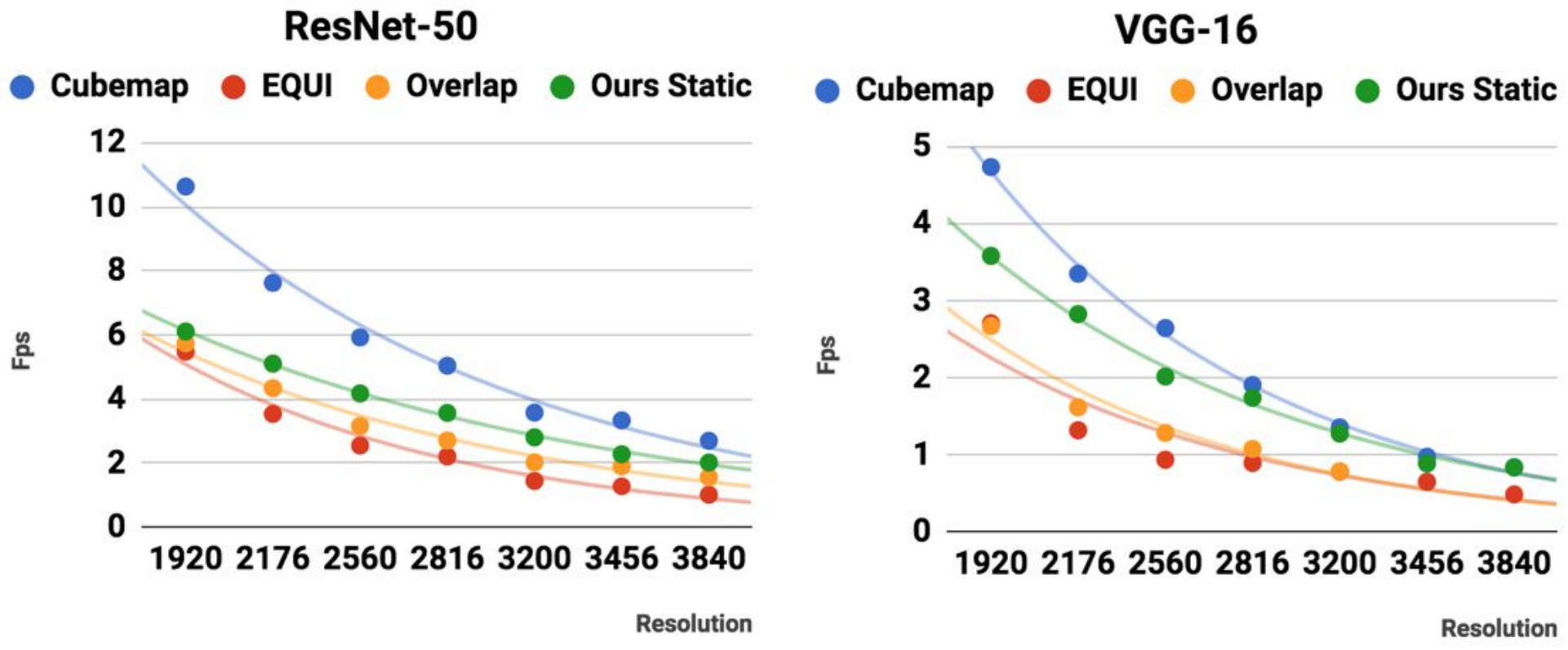}
\end{center}\vspace{-4mm}
\caption{\small Speed of static methods. h-axis represents image resolution, v-axis represents FPS. As the resolution increase, the speed of Ours Static becomes closer to Cubemap. Besides, Ours Static exceeds EQUI and Overlap in FPS for all the tested resolutions.}
\label{fig.statspeed}
\end{figure}

\begin{figure}[t!]\vspace{-4mm}
\begin{center}
\includegraphics[width=1\linewidth]{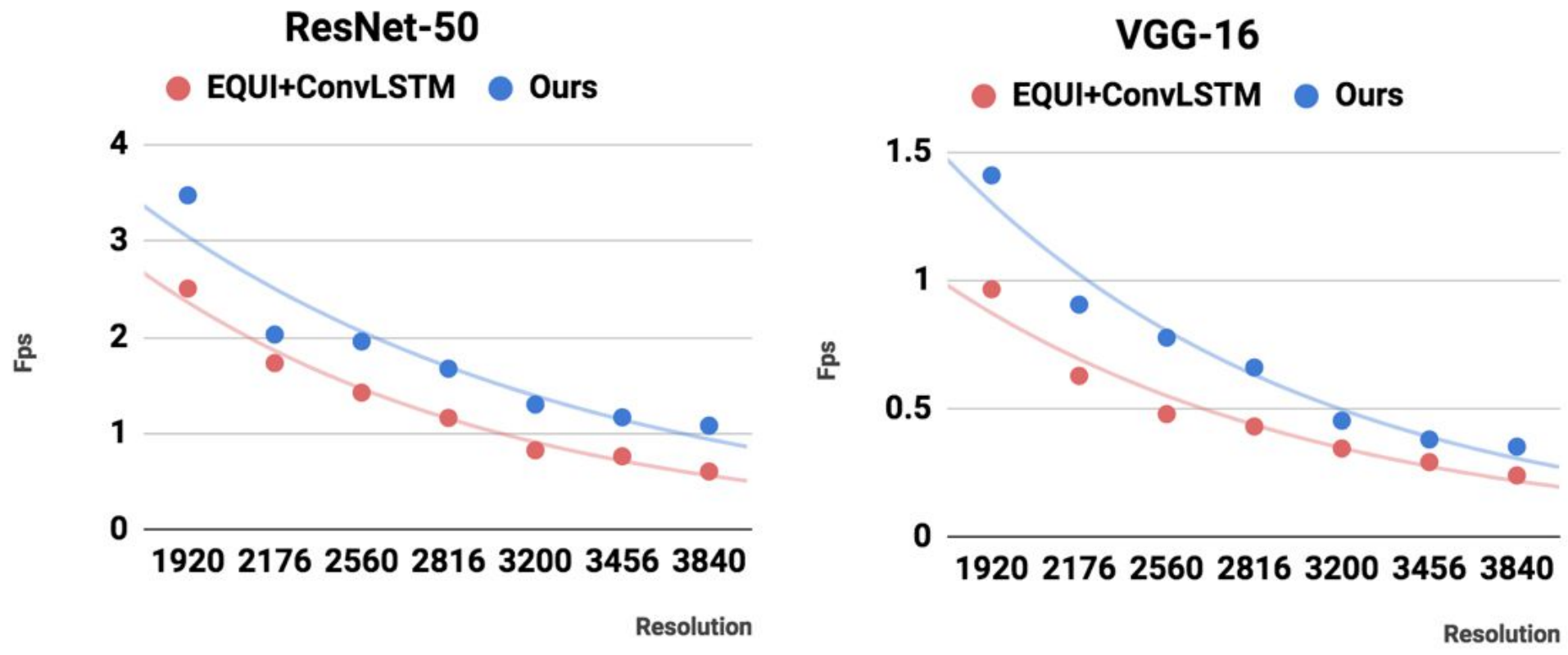}
\end{center}\vspace{-4mm}
\caption{\small Speed of temporal methods. h-axis represents image resolution, v-axis represents FPS. Ours is faster than EQUI + ConvLSTM.}\label{fig.tempspeed}
\end{figure}

To compare the inference speed of our approach with other baselines with common resolution scale of 360$^\circ$ videos, we conduct an experiment to measure the Frame-Per-Second (FPS) along different resolutions. Fig.~\ref{fig.statspeed} shows the speed of static methods including Cubemap, EQUI, Overlap, and our static model (Ours Static). Fig.~\ref{fig.tempspeed} shows the speed comparison between two methods using ConvLSTM: EQUI+ConvLSTM and our temporal model (Ours). The left and right side of both figures is for ResNet-50 and VGG-16, respectively. The resolutions are set from 1920 (Full HD) to 3840 (4K).
The result of Fig~\ref{fig.statspeed} shows that Ours Static is slower than Cubemap but faster than Overlap and EQUI. Note that at the same amount of time, Ours Static has the ability to compute with a frame much larger than EQUI. Additionally, Fig.~\ref{fig.tempspeed} shows that Ours is significantly faster than EQUI+ConvLSTM. We evaluate the computational efficiency on NVIDIA Tesla M40 GPU.


\vspace{-2mm}
\subsection{Evaluation metrics}\label{sec.eval_metrics}
\vspace{-3mm}

We refer to the MIT Saliency Benchmark \cite{mit-saliency-benchmark} and report three common metrics:

\noindent{\textbf{AUC-Judd (AUC-J).}}
AUC-Judd \cite{Judd12} measures differences between our saliency prediction and the human labeled ground truth by calculating the true positive and false positive rate for the viewpoints.

\noindent{\textbf{AUC-Borji (AUC-B).}}
AUC-Borji score uniformly and randomly samples image pixels as negative and defines the saliency map values above the threshold at these pixels as false positives.

\noindent{\textbf{Linear Correlation Coefficient (CC).}}
The linear correlation coefficient is a distribution based metric to measure the linear relationship of given saliency maps and the viewpoints. The coefficient value is bounded between -1 and 1, representing the linear dependencies between our output and ground truth.

\begin{figure*}[t!]\vspace{-5mm}
\begin{center}
\includegraphics[width=1\linewidth]{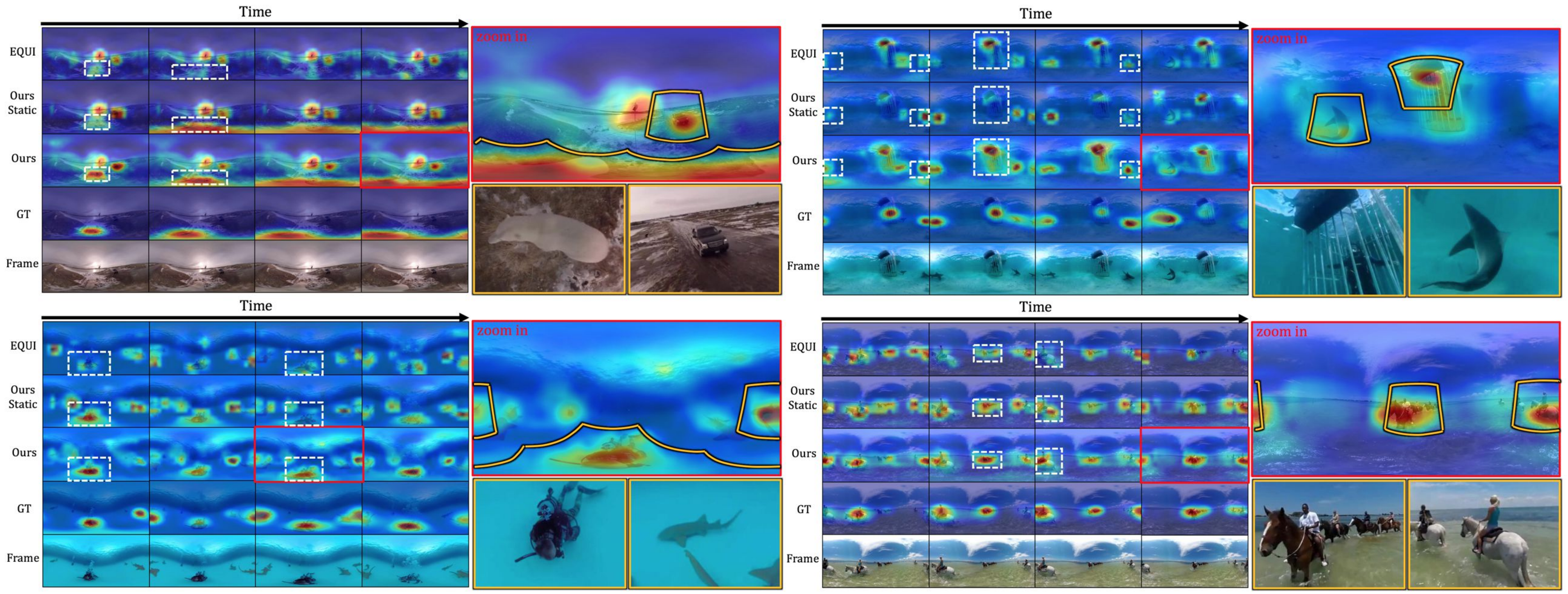}
\end{center}\vspace{-4mm}
\caption{\small Qualitative examples. In each block, consecutive frames of various methods, ground truth, and raw videos are shown in the left panel. We highlight regions for comparison using white dash rectangles. In the right panel, one example is zoom-in (red box) and two salient NFoVs (yellow boxes) are rendered. Our temporal method (Ours) significantly outperforms others in overcoming distortion, image boundaries, and smoothness in time. See more examples in supplementary materials.
}
\label{fig.qual}
\end{figure*}

\begin{figure*}[t!]\vspace{-2mm}
\begin{center}
\includegraphics[width=0.975\linewidth]{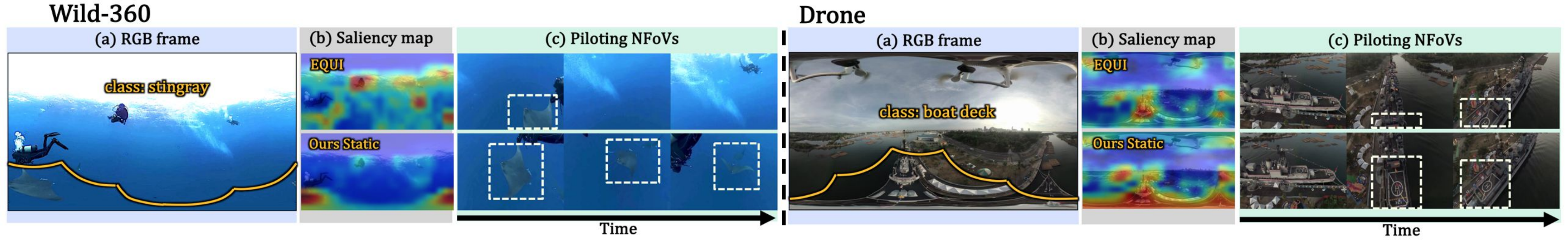}
\end{center}\vspace{-4mm}
\caption{\small NFoV piloting results. The left and right sides show the result of Wild-360 dataset and Drone data, respectively. Panel (a) shows a video frame with targeted FoV drawn in orange. Panel (b) shows EQUI (top) and Ours Static (bottom) saliency maps. Panel (c) shows NFoVs in multiple time steps with white dash boxes indicating the targeted regions. NFoV Piloting with Ours Static in all 4 videos is more capable to capture the target regions.}
\label{fig.pilot}
\end{figure*}

\vspace{-2mm}
\subsection{Saliency comparison}\label{sec.result}
\vspace{-2mm}

\small
\begin{table}[t!]
\centering
\begin{tabular}{|l|l|l|l|}
\hline
\textit{VGG-16}      & CC   & AUC-J & AUC-B \\ \hline \hline
Cubemap     & 0.338	& 0.797	& 0.757      \\ \hline
Overlap    & 0.380	&0.836	& 0.813   \\ \hline
EQUI        & 0.285	& 0.714 & 0.687     \\ \hline
EQUI + ConvLSTM & 0.330	& 0.823	& 0.771   \\ \hline \hline
\textbf{Ours Static}    & 0.381	& 0.825	& 0.797    \\ \hline
\textbf{Ours}   &\textbf{0.383}	&\textbf{0.863}	&\textbf{0.843}     \\ \hline \hline
\hline
\textit{ResNet-50}       & CC   & AUC-J & AUC-B \\ \hline \hline
Cubemap      & 0.413	&0.855	&0.836    \\ \hline
Overlap      &0.383	&0.845	&0.825    \\ \hline
EQUI         & 0.331 & 0.778 & 0.741    \\ \hline
EQUI + ConvLSTM &  0.337	& 0.839	& 0.783   \\ \hline \hline
\textbf{Ours Static}    & \textbf{0.448}	& 0.881	& 0.852    \\ \hline 
\textbf{Ours}   &0.420	&\textbf{0.898}	&\textbf{0.859}     \\ \hline
\hline \hline
\textit{Baselines}           & CC   & AUC-J & AUC-B \\ \hline \hline
Motion Magnitude \cite{weinzaepfel2013deepflow}   & 0.288 &0.687	&0.642    \\ \hline
ConsistentVideoSal \cite{wang2015consistent}      &  0.085 &0.547 &0.532  \\ \hline
SalGAN \cite{pan2017salgan}    &0.312 &0.717 &0.692     \\ \hline
\end{tabular}
\caption{\small Saliency prediction accuracy in CC, AUC-J, and AUC-B. Our methods (bold font) consistently outperform all baseline methods. See Sec.~\ref{sec.base} for the baseline methods compared.}\label{table.saliency}
\end{table}
\normalsize
From saliency comparison shown in Table.~\ref{table.saliency}, we observe the following: (1) \textit{Our temporal model is the best} in all evaluation metrics except one where our static model is better; (2) \textit{ConvLSTM improves performance} since typically EQUI+ConvLSTM outperforms EQUI and Ours outperforms Ours Static.
Typical examples are shown in Fig.~\ref{fig.qual}, where we compare EQUI, Ours Static, with Ours. Our temporal model typical predicts smooth saliency map in time and is more effective to salient regions on image boundaries or in the top/bottom part of the image.

\vspace{-2mm}
\subsection{NFoV piloting}\label{sec.result}
\vspace{-3mm}
We use Wild-360 dataset and a set of Drone videos to demonstrate our result.
Our scenario is to generate class specific NFoV trajectory. 
We use our per-frame saliency feature $\mathcal{M}$ to get the score of each sampled viewing angle by average all the scores inside its corresponding NFoV on $\mathcal{M}$. Then $\mathcal{O}$ is extracted by $\mathcal{M}(\hat{c})$, where $\hat{c}$ is a class index decided by user. To link per-frame NFoV's to a trajectory by saliency score, we use AUTOCAM~\cite{supano2vid} to find a feasible path of salient viewpoints. Fig.~\ref{fig.pilot} shows that the NFoV tracks we generated are able to capture salient viewpoints better than equirectangular.

\begin{table}[t!]
\centering
\begin{tabular}{|l|l|l|}
\hline
Methods & win / loss\\ \hline
Ours Static vs. EQUI & 95 / 65\\ \hline
Ours Static vs. Cubemap & 97 / 63\\ \hline
Ours vs. Ours Static & 134 / 26\\ \hline
Ours vs. GT & 70 / 90\\ \hline
\end{tabular}\vspace{1mm}
\caption{\small Human evaluation results. We have 16 viewers watching 10 clips in each row so the number of wins and losses combined is 160. The result shows that Ours Static outperforms other static baselines; Ours outperforms Ours Static and is comparable to GT.}
\label{table.humaneval_sal}
\end{table}


\vspace{-2mm}
\subsection{Human evaluation}\label{sec.humaneval}
\vspace{-3mm}
We design 4 user tests and compare our methods with different image formats (Ours Statics vs. EQUI and Our Static vs. Cubemap) and the performance of our temporal model (Ours vs. Ours Static and Ours vs. GT). Settings of the 4 tests are shown in Table.~\ref{table.humaneval_sal}. In each test, we pick 10 different clips from the Wild-360 test set and generate saliency map prediction from 2 different methods. We ask 16 viewers to select the saliency map prediction which (1) activates on salient regions more correctly, (2) is smoother across frames. We further conduct a two-tailed binomial test and it shows that Ours Static is statistically superior to EQUI and Cubemap with p-value $<$ 0.05. This implies that Ours Static is a strong preference over using EQUI and Cubemap. Moreover, Ours significantly outperforms Ours Static with p-value $<$ 0.01. When compared with ground truth, the p-value is 0.13 which implies that the saliency map generated by Ours is comparable with ground truth.

\vspace{-5mm}
\section{Conclusion}\label{sec.Con}
\vspace{-3mm}
We propose a novel spatial-temporal network which is (1) weakly-supervised trained without 360$^\circ$ supervision and (2) tailor-made for 360$^{\circ}$ viewing sphere, where a simple and effective Cube Padding (CP) technique is introduced. On a newly collected Wild-360 dataset with challenging videos and saliency heatmap annotations, our method outperforms state-of-the-art methods in both speed and quality.
\cutsectionup
\vspace{-2mm}
\section*{Acknowledgements}\label{sec.Ack} 
\vspace{-3mm}
We thank MOST-106-3114-E-007-008 and MediaTek for their support.

{\small
\bibliographystyle{ieee}
\bibliography{egbib.bib}
}

\end{document}